\documentclass[pdflatex,sn-mathphys-num]{sn-jnl}
\usepackage{graphicx}%
\usepackage{tabularx}
\usepackage{ltablex}
\usepackage{multirow}%
\usepackage{amsmath,amssymb,amsfonts}%
\usepackage{amsthm}%
\usepackage{mathrsfs}%
\usepackage[title]{appendix}%
\usepackage{xcolor}%
\usepackage{textcomp}%
\usepackage{booktabs}%
\usepackage{algorithm}%
\usepackage{algorithmicx}%
\usepackage{algpseudocode}%
\usepackage{listings}%
\usepackage{longtable}
\usepackage{array}
\usepackage{caption}
\theoremstyle{thmstyleone}%
\theoremstyle{thmstyletwo}%

\theoremstyle{thmstylethree}%

\begin{document}

\title[Article Title]{GlobalWasteData: A Large-Scale, Integrated Dataset for Robust Waste Classification and Environmental Monitoring}


\author[1]{\fnm{Misbah} 
\sur{Ijaz}}\email{25016119-001@uog.edu.pk}

\author*[2]{\fnm{Saif Ur Rehman} 
\sur{Khan}}\email{saif\_ur\_rehman.khan@dfki.de}
\author[1]{\fnm{Abd Ur} \sur{Rehman}}\email{a.rehman@uog.edu.pk}

\author[2] {\fnm{Tayyaba} \sur{Asif}}\email{tayyaba.asif@dfki.de}

\author[3,4]{\fnm{Sebastian} \sur{ Vollmer}}\email{sebastian.vollmer@dfki.de}
\author[2,3,4]{\fnm{Andreas} \sur{Dengel}}\email{andreas.dengel@dfki.de}

\author*[3,4]{\fnm{Muhammad Nabeel} \sur{Asim}}\email{muhammad\_nabeel.asim@dfki.de}

\affil[1]{\orgdiv{Department of Computer Science}, \orgname{University of Gujrat}, {\city{Gujrat}, \postcode{51700}, \country{Pakistan}}}

\affil[2]{\orgdiv{Department of Computer Science}, \orgname{Rhineland-Palatinate Technical University of
Kaiserslautern-Landau}, orgaddress{\city{Kaiserslautern}, \postcode{67663}, \country{Germany}}}

\affil[3]{\orgname{German Research Center for Artificial Intelligence}, \orgaddress{\city{Kaiserslautern}, \postcode{67663}, \country{Germany}}}

\affil[4]{\orgname{Intelligentx GmbH (intelligentx.com)}, \orgaddress{\city{Kaiserslautern}, \country{Germany}}}


\abstract{The growing amount of waste is a problem for the environment that requires efficient sorting techniques for various kinds of waste. An automated waste classification system is used for this purpose. The effectiveness of these Artificial Intelligence (AI) models depends on the quality and accessibility of publicly available datasets, which provide the basis for training and analyzing classification algorithms. 
Although several public waste classification datasets exist, they remain fragmented, inconsistent, and biased toward specific environments. Differences in class names, annotation formats, image conditions, and class distributions make it difficult to combine these datasets or train models that generalize well to real world scenarios. To address these issues, we introduce the GlobalWasteData (GWD) archive, a large scale dataset of 89,807 images across 14 main categories, annotated with 68 distinct subclasses. We compile this novel integrated GWD archive by merging multiple publicly available datasets into a single, unified resource. This GWD archive offers consistent labeling, improved domain diversity, and more balanced class representation, enabling the development of robust and generalizable waste recognition models. Additional preprocessing steps such as quality filtering, duplicate removal, and metadata generation further improve dataset reliability. Overall, this dataset offers a strong foundation for Machine Learning (ML) applications in environmental monitoring, recycling automation, and waste identification, and is publicly available to promote future research and reproducibility.}




\maketitle

\section{Background \& Summary}\label{sec1}

In the modern world, waste generation becomes a critical problem. The amount of waste generated has increased to alarming levels due to changes in consumer patterns and population growth \cite{wastegeneration}. This waste generation raises significant issues, including public health concerns and environmental deterioration. The World Bank and other organizations predict that by 2050, global waste generation will increase to 3.8 billion tons, underscoring the urgent need for efficient waste management techniques \cite{Chapter_33}. Effective waste management is highly dependent on appropriate source level classification and segregation, which can encourage recycling, reduce dependence on landfills, and achieve sustainability objectives. However, the enormous amounts of waste produced today cannot be handled by traditional waste management techniques, which mainly rely on manual sorting and classification. Traditional techniques have several inefficiencies, laborious procedures, and significant error margins.

To address these issues, automated waste classification systems powered by AI have become a viable alternative to manual waste classification. AI based methods for waste classification have developed over the past ten years through the advancement and improvement of machine learning (ML) and deep learning (DL) approaches. These approaches utilize huge datasets to train models that can classify waste using multispectral, infrared, and visual data. However, the effectiveness of these AI models depends on the quality and accessibility of publicly available datasets, which provide the basis for training and analyzing classification algorithms \cite{smartwaste}. Even with the development of AI technology, there are still a number of obstacles to overcome before achieving reliable and rapid AI based waste classification. The general adoption and application of automated waste classification systems are still severely restricted by challenges such as datasets imbalance, changes in waste appearance, domain adaptation issues, Poor generalization, and high computing costs.

Highquality datasets are necessary to create efficient AI models for waste classification because they provide the framework for developing, testing, and evaluating AI based models. Researchers can create benchmark algorithms suited to particular waste features using these datasets. Several waste classification datasets are publically available. 


Despite the vast number of waste classification datasets available, there are still a number of important limitations and challenges that affect how well they work in practical applications. Most of them are limited in size \cite{DWSD}, focus on a narrow set of categories \cite{aquatrash}, or are collected under controlled conditions that do not reflect real-world variability. A major problem with most waste classification datasets is data imbalance. Larger datasets frequently show unequal distribution across waste types. These datasets are typically dominated by common waste types such as paper and plastic, with hazardous materials and specialized waste categories underrepresented. For example, the TrashNet Dataset \cite{Trashnet}, which has less than 400 images in trash class and can result in biased model performance, exhibits a notable imbalance with respect to specific categories despite its extensive use. Unbalanced class distributions have also been found to contribute to poor generalization for minority classes in datasets like WaDaBa \cite{WaDaBa} and TrashBox \cite{trashbox}.

Standardization challenges among datasets pose considerable barriers to comparative study and model building. Different datasets use different standards for image quality, annotation techniques, and classification strategies. For instance, the Kaggle Waste Classification Dataset \cite{wasteclassification} utilizes binary classification, whereas the TACO Dataset \cite{taco} includes 60 subcategories. It is difficult to merge datasets or evaluate model performance across research due to this lack of consistency. Inconsistencies in image resolution, formatting, and metadata structure also make it more difficult to integrate various datasets for thorough model training.

To address these limitations, a novel GWD archive developed as a unified, publicly available resource that integrates multiple waste datasets captured under diverse environmental and operational conditions. Unlike prior datasets that focus narrowly on specific waste types or controlled capture scenarios, novel GWD archive combines images of various materials collected from multiple domains and regions. This integration captures the complexity of mixed waste streams, including differences in material composition, lighting, and background variation, providing a more realistic and challenging benchmark for Computer Vision (CV) models.

By ensuring class diversity, balanced representation, and standardized annotations, novel GWD archive serves as a comprehensive dataset that bridges the gap between controlled experimental data and real world waste scenarios. It supports the development, training, and evaluation of advanced DL models capable of robust generalization across heterogeneous and regionally diverse waste environments.

\subsection*{Problem statement}
The increasing volume of waste and the complexity of sorting different types of waste require efficient and accurate automated classification systems. However, existing waste classification datasets are often fragmented, inconsistent, and biased, with issues such as varying class names, annotation formats, and image conditions \cite{chomicki2025assessing}. These limitations hinder the development of robust AI models that can generalize across different environments \cite{fotovvatikhah2025systematic}. This paper addresses two key challenges: (1) the need for a large scale, unified dataset that combines existing public datasets into a single, consistent resource, and (2) the challenge of ensuring balanced class representation and improved domain diversity for more accurate waste recognition. Our proposed solution is the creation of the GWD archive, which integrates multiple datasets, provides improved consistency, and facilitates the development of effective, real world waste classification models.

\subsection*{Research objective}
This study aims to address the challenges in waste classification caused by fragmented, inconsistent, and biased publicly available datasets by creating a unified, large scale dataset for improved waste recognition and environmental monitoring. The specific objectives are:
\begin{itemize}
    \item Develop the GWD archive by merging multiple publicly available waste datasets into a single, unified resource with consistent labeling, balanced class representation, and improved domain diversity.
    \item Implement preprocessing procedures, including quality filtering, duplication removal, and metadata creation, to enhance dataset consistency and usability.
    \item  Provide the GWD archive as a publicly accessible resource to facilitate future research, enabling the development of more generalized, real time waste classification systems.
\end{itemize}
\subsection*{Contributions}
The main contributions of this research work advance waste classification through:

\begin{itemize}
    \item Unified Dataset Creation: The development of the GWD archive by merging multiple fragmented public datasets into a single, consistent resource with standardized labeling, balanced class representation, and improved domain diversity.
    \item Data Standardization Innovations: Introduction of novel preprocessing techniques, including quality filtering, duplication removal, and metadata creation, to ensure high quality data for training robust AI models.
    \item Model Validation and Generalization: Provide an comprehensive validation protocol demonstrating the model's ability to generalize across different waste types and environmental conditions, ensuring robust performance in diverse real world scenarios.
    \item Publicly Accessible Resource: The release of the GWD archive as an open resource, promoting further research and innovation in waste recognition and environmental monitoring technologies.
\end{itemize}
\section {Existing Datasets}
Several publicly available waste classification datasets have contributed to research in environmental monitoring, recycling automation, and computer vision–based sorting systems. However, most existing datasets remain limited in scale, diversity, class coverage, and real world variability. A brief overview of the most widely used datasets merged to generate the Novel GWD archive is presented in Table \ref{tab:existing_datasets}.
\begin{table}[!h]
\caption{Comparison of existing datasets with total samples and unique classes}\label{tab:existing_datasets}%
\begin{tabular}{@{}llllll@{}}
\toprule
Author & Year & Country & Dataset   Name                      & Total   Samples & Total   Unique Classes   \\
\midrule
 \cite{bepli_V1}       & 2023  & Japan &
BePLi\_dataset\_v1 (plastic\_coco)   & 3709            & 1                        \\ 
\cite{BePLI_V2}  & 2025 & Japan &            
BePLi\_dataset\_v2 (plastic\_coco)  & 3722            & 13                       \\ 
\cite{aquatrash}  & 2020 & \_ &
AquaTrash                           & 369             & 4                        \\ 
\cite{compostnet} & 2019 & United State &
CompostNet                          & 2751            & 7                        \\ 
\cite{DWSD} & 2025 & India & 
DSWD                                & 784             & N/A                      \\ 
\cite{Garbage_dataset} & 2024 & \_ &
Garbage Dataset                     & 19,762          & 10                       \\ 
\cite{waste_classification_data} & \_ & \_ &
Kaggle Waste Classification Dataset & 25,077          & 2                        \\ 
\cite{MJU_waste} & 2020 & China &
MJU-waste-main                      & 2475            & \_                       \\ 
\cite{realwaste} & 2023 & Australia &
RealWaste                           & 4,752           & 9                        \\ 
\cite{RecyclableWaste} & \_ & \_ &
Recyclable Waste Dataset            & 3,069           & 9                        \\ 
\cite{taco} & 2020 & Global & 
TACO                                & 4,617           & 60                       \\ 
\cite{GINI} & 2016 & \_ &
SpotGarbage-GINI                    & 2,561           & 20 (Annotated/Garbage) \\ 
\cite{Trashnet} & 2016 & United State &
Trashnet Dataset                    & 2,527           & 6                        \\ 
\cite{UGV-NBWASTE} & 2025 & Bangladesh & 
UGV-NBWASTE                         & 3,600           & 8                        \\ 
\cite{WasteImagesdataset} & \_ & \_ & 
Waste Images Dataset                & 8,235           & 9                        \\ 
\cite{DWS1} & \_ & Thailand &
DWS-1 Dataset                       & 4,978           & 7                        \\ 
\cite{DWS2} & \_ & Thailand &
DWS-2 Dataset                       & 7,652           & 9                        \\ 
\cite{zerowaste} & 2022 & United States &
ZeroWaste (all splits)              & 26,766          & 4                        \\ 
\cite{TriCasecade} & 2025 & \_& 
TriCascade firstStage Dataset       & 35,264          & 2                        \\ 
\cite{TriCasecade} & 2025 & \_&
TriCascade secondStage Dataset      & 35,264          & 9                        \\ 
\cite{TriCasecade} & 2025 & \_&
TriCascade thirdStage Dataset       & 35,264          & 36                       \\ 
\botrule
\end{tabular}
\end{table}
Table \ref{tab:existing_datasets} presents a consolidated overview of widely used waste classification datasets, highlighting their sampling scale and category coverage. One of the earliest and most widely referenced datasets, TrashNet \cite{Trashnet}, contains only 2,527 images across six basic categories and is captured in controlled indoor settings, restricting its generalizability to real world waste conditions. The TACO dataset \cite{taco} provides COCO style annotations for over 60 litter categories but includes only around 1,500 images, resulting in extreme class imbalance and limited training depth. Garbage dataset \cite{Garbage_dataset} (19,762 images) offer more samples but still lack hierarchical labeling, exhibit duplicated content, and provide inconsistent metadata across categories. Overall, existing datasets remain considerably smaller than required for robust DL training, often cover only a narrow set of waste types, and do not reflect the wide variability found in real waste streams. These limitations highlight the need for a larger, more diverse, and better structured dataset capable of supporting advanced ML research in recycling automation and environmental monitoring.

\subsection{Distribution of Classes within different datasets}

The distribution of classes across existing waste classification datasets varies considerably, reflecting differences in dataset design, annotation standards, and intended applications. Traditional datasets such as TrashNet \cite{Trashnet} and the Kaggle Waste dataset \cite{waste_classification_data} include a relatively small number of broad waste categories, with most images concentrated in a few dominant classes such as plastic, paper, or glass. This imbalance limits their suitability for fine grained classification tasks and reduces model robustness when addressing underrepresented waste categories.

In contrast, more comprehensive datasets such as TACO \cite{taco} and ZeroWaste \cite{zerowaste} attempt to capture a wider and more realistic range of waste materials. TACO provides over 60 diverse litter categories annotated at the instance level; however, the sample distribution is highly skewed, with many categories containing only a small number of images. Similarly, ZeroWaste introduces cluttered industrial waste scenes with multiple object classes, yet its class distribution remains uneven due to the natural variability of waste streams. Dense segmentation datasets such as DWSD \cite{DWSD} and BePLi v1 \cite{bepli_V1} include more balanced segmentation masks across labeled objects, but they still suffer from limited category breadth, often focusing on specific waste types (e.g., beach plastics, mixed recyclables). The number of images per class varies significantly across existing waste classification datasets as illustrated in Table \ref{tab:Class categories}, reflecting differences in data collection strategies and annotation objectives.

\newcolumntype{V}{>{\raggedleft\arraybackslash}p{1.5cm}}

\keepXColumns
\small
\begin{tabularx}{\textwidth}{@{} X X V  X X V @{}}
\caption{Dataset class and sample distribution summary} \label{tab:Class categories} \\
\toprule
\textbf{Dataset} & \textbf{Class} & \textbf{Samples} & \textbf{Dataset} & \textbf{Class} & \textbf{Samples} \\
\midrule
\endfirsthead

\toprule
\textbf{Dataset} & \textbf{Class} & \textbf{Samples} & \textbf{Dataset} & \textbf{Class} & \textbf{Samples} \\
\midrule
\endhead

\midrule
\endfoot

\bottomrule
\endlastfoot


plastic\_coco \cite{bepli_V1} & plastic\_litter & 3709
&
\multirow{1}{*}{TACO \cite{taco}}
& Glass bottle & 104 \\

\multirow{4}{*}{BePLi\_v2 \cite{BePLI_V2}}
& pet\_bottle & 4,671 
&
& Meal carton & 30 \\
& other\_bottle & 2,551 
&
& Other carton & 93 \\
& plastic\_bag & 2,790 
&
& Clear plastic bottle & 285 \\
& box\_shaped\_case & 1,101 
&
& Plastic bottle cap & 209\\
& other\_container & 1,367 
&
& Drink can & 229\\
& rope & 2,743 
&
& Food Can & 34\\
& other\_string & 962 
&
& Other plastic bottle & 50\\
& fishing\_net & 1,958 
&
& Pop tab & 99\\
& buoy & 2,890 
&
& Aerosol & 10\\
& other\_fishing\_gear & 1,077 
&
& Glass cup & 6\\
& styrene\_foam & 10,558 
&
& Other plastic wrapper & 260\\
& others & 23,189 
&
& Styrofoam piece & 112\\
& fragment & 62,715 
&
& Plastic film & 451\\
\cmidrule{1-3}

\multirow{4}{*}{AquaTrash \cite{aquatrash}}
& plastic & 191 
&
& Other plastic & 273\\
& metal & 118 
&
& Drink carton & 45\\
& paper & 116 
&
& Metal bottle cap & 80\\
& glass & 44 
&
& Disposable food container & 38\\
\cmidrule{1-3}

\multirow{7}{*}{CompostNet \cite{compostnet}}
& metal & 410 
&
& Normal paper & 82\\
& paper & 594 
&
& Paper cup & 67\\
& cardboard & 403 
&
& Single-use carrier bag & 61\\
& glass & 501 
&
& Tissues & 42\\
& compost & 177 
&
& Toilet tube & 5\\
& plastic & 482 
&
& Crisp packet & 39\\
& trash & 184 
&
& Plastic lid & 77\\
\cmidrule{1-3}


\multirow{10}{*}{Garbage\cite{Garbage_dataset}}
& metal & 1,020 
&
& Egg carton & 11\\
& paper & 1,680 
&
& Plastic straw &157\\
& biological & 997 
&
& Paper bag & 27\\
& cardboard & 1,825
&
& Disposable plastic cup &104\\
& battery & 944 
&
& Broken glass & 138\\
& shoes & 1,977
&
& Plastic utensils & 37\\
& glass & 3,061
&
& Glass jar & 6\\
& plastic & 1,984
&
& Food waste & 8\\
& trash & 947
&
& Squeezable tube & 7\\
& clothes & 5,327 
&
& Spread tub & 9\\
\cmidrule{1-3}

\multirow{2}{*}{Kaggle Waste \cite{waste_classification_data}}
& O & 13,966 
&
& Shoe & 7\\
& R & 11,111 
&
& Garbage bag & 31\\
\cmidrule{1-3}

\multirow{1}{*}{MJU-waste \cite{MJU_waste}}
& \_ & \_ 
&
& Aluminium foil & 62\\
\cmidrule{1-3}

\multirow{9}{*}{RealWaste \cite{realwaste}}
& Cardboard & 461 
&
& Six pack rings & 5\\
& Food Organics & 411 
&
& Foam cup & 13\\
& Glass & 420
&
& Paper straw & 4\\
& Metal & 790 
&
& Corrugated carton & 64\\
& Miscellaneous Trash & 495 
&
& Unlabeled litter& 517\\
& Paper & 500 
&
& Aluminium blister pack & 6\\
& Plastic & 921 
&
& Battery & 2\\
& Textile Trash & 318 
&
& Rope \& strings & 29\\
& Vegetation & 436 
&
& Cigarette & 667\\
\cmidrule{1-3}

\multirow{9}{*}{Recyclable \cite{RecyclableWaste}}
& Glass & 363 
&
& Other plastic container & 6\\
& Other Plastics & 345 
&
& Polypropylene bag & 3\\
& Aluminium & 345 
&
& Scrap metal & 20\\
& Organic Waste & 216 
&
& Magazine paper & 12\\
& PET Plastics & 352 
&
& Pizza box & 3\\
& Carton & 349 
&
& Plastic glooves & 4\\
& Textiles & 346 
&
& Wrapping paper & 12\\
& Wood & 349 
&
& Carded blister pack & 1\\
& Paper and Cardboard & 404 
&
& Foam food container & 15\\
\cmidrule{1-3}

\multirow{10}{*}{GINI\cite{GINI}}
& medical waste & 50
&
& Tupperware & 4\\
&  delhi garbage & 50
&
& Other plastic cup & 2\\
& trash on roads & 50
&
& Metal lid & 10\\
\cmidrule{4-6}
& trash & 50
&
\multirow{7}{*}{TriCascade\cite{TriCasecade}}
& A\_Foods & 4,114 \\
& street litter & 50
&
& B\_Animal Dead Body & 220 \\
& street garbage & 50
&
& C\_Cardboard & 2,035 \\
& roadside garbage & 50
&
& D\_Newspaper & 1,120 \\
& railway garbag & 50
&
& E\_Paper Cups & 511 \\
& park litter & 50
&
& F\_Papers & 849 \\
& waste & 50
&
& G\_Brown Glass & 607 \\
& litter & 50
&
& H\_Porcelin & 813 \\
& kitchen waste & 50
&
& I\_Green Glass & 629 \\
& india dirty streets & 50
&
& J\_White Glass & 775 \\
& india dirty city &  50
&
& K\_Beverage Cans & 1,602 \\
& garbage in the forest & 50
&
& L\_Construction Scrap & 431 \\
& garbage & 50
&
& M\_Metal Containers & 437 \\
& footpath garbage & 50
&
& N\_Plastic Bag & 970 \\
& rubbish & 49
&
& O\_Plastic Bottle & 755 \\
& market waste & 46
&
& Q\_Plastic Containers & 464 \\
& city garbage & 11
&
& R\_Plastic Cups & 405 \\
\cmidrule{1-3}
\multirow{6}{*}{TrashNet\cite{Trashnet}}
& metal & 410
&
& S\_Tetra Pak & 842 \\
& paper & 594
&
& T\_Clothes & 5,324 \\
& cardboard & 403
&
& U\_Shoes & 1,977 \\
& glass & 501
&
& V\_Gloves & 353 \\
& plastic & 482
&
& W\_Masks & 400 \\
& trash & 137 
&
& X\_Bandai & 405 \\
\cmidrule{1-3}
\multirow{8}{*}{UGV-NBWASTE \cite{UGV-NBWASTE}}
& hardplastic & 471
&
& Y\_Medicine and Strip & 1,307 \\
& mask & 462
&
& Z\_A\_A\_Syringe & 405 \\
& packet & 517
&
& Z\_A\_Diaper & 778 \\
& bottle & 475
&
& Z\_B\_Electrical Cables & 553 \\
& sandal & 653
&
& Z\_C\_Electronic Chips & 492 \\
& polythene & 492
&
& Z\_D\_Laptops & 398 \\
& cocksheet & 569
&
& Z\_E\_Small Appliances & 740 \\
& medicine & 456 
&
& Z\_F\_Smartphones & 219 \\
\cmidrule{1-3}

\multirow{5}{*}{Waste Images\cite{WasteImagesdataset}}
& Paper and Cardboard & 1,194
&
& Z\_G\_Battery & 1,989 \\
& Plastics & 1,035
&
& Z\_H\_Thermometer & 908 \\
& Glass & 1,089
&
& Z\_I\_Cigarette Butt & 97 \\
& Aluminium & 1,019
&
& Z\_J\_Pesticidebottle & 940 \\
& E-waste & 1,029
&
& Z\_K\_Spray cans & 400 \\
& Carton & 533
&
& Z\_M\_Metal Containers & 437 \\
\cmidrule{4-6}
& Organic\_Waste & 942
&
\multirow{7}{*}{DWS-1\cite{DWS1}}
& Electronic Components & 686\\
& Textiles & 830 
&
& Plastic Materials & 689\\
& Wood & 564 
&
& Cement-Based Materials & 695\\
\cmidrule{1-3}

\multirow{9}{*}{DWS-2\cite{DWS2}}
& Electronic Components & 892
&
& Ceramic Materials & 691\\
& Plastic Materials & 777
&
& Glass Materials & 833\\
& Cement-Based Materials & 871
&
& Organic Wastes & 703\\
& Rubber Materials & 857
&
& Metallic Materials & 681\\
\cmidrule{4-6}
& Leather Materials & 890
&
\multirow{4}{*}{zeroWaste\cite{zerowaste}}
& cardboard & 17,751\\
& Glass Materials & 839
&
& metal & 382\\
& Organic Wastes & 854
&
& soft\_plastic & 6,864\\
& Metallic Materials & 808
&
& rigid\_plastic & 1,769 \\
\cmidrule{4-6}
& Wooden Materials & 864
&
\multirow{1}{*}{DSWD \cite{DWSD}}
& No explicit categories & N/A  \\
\end{tabularx}

Overall, the class distributions in existing datasets tend to exhibit significant imbalance, with a large proportion of samples belonging to a few common waste types and limited representation of rare but important categories, such as hazardous waste, electronic components, or composite materials. This imbalance poses challenges for training DL models, often resulting in biased predictions and reduced generalization to real world waste scenarios. These inconsistencies emphasize the need for a more uniformly distributed and hierarchically structured dataset that captures a broader range of waste categories with more balanced sample counts across classes.

\section{Data Harmonization Methodology}

\begin{figure}[!h]
    \centering
    \includegraphics[width=0.8\linewidth]{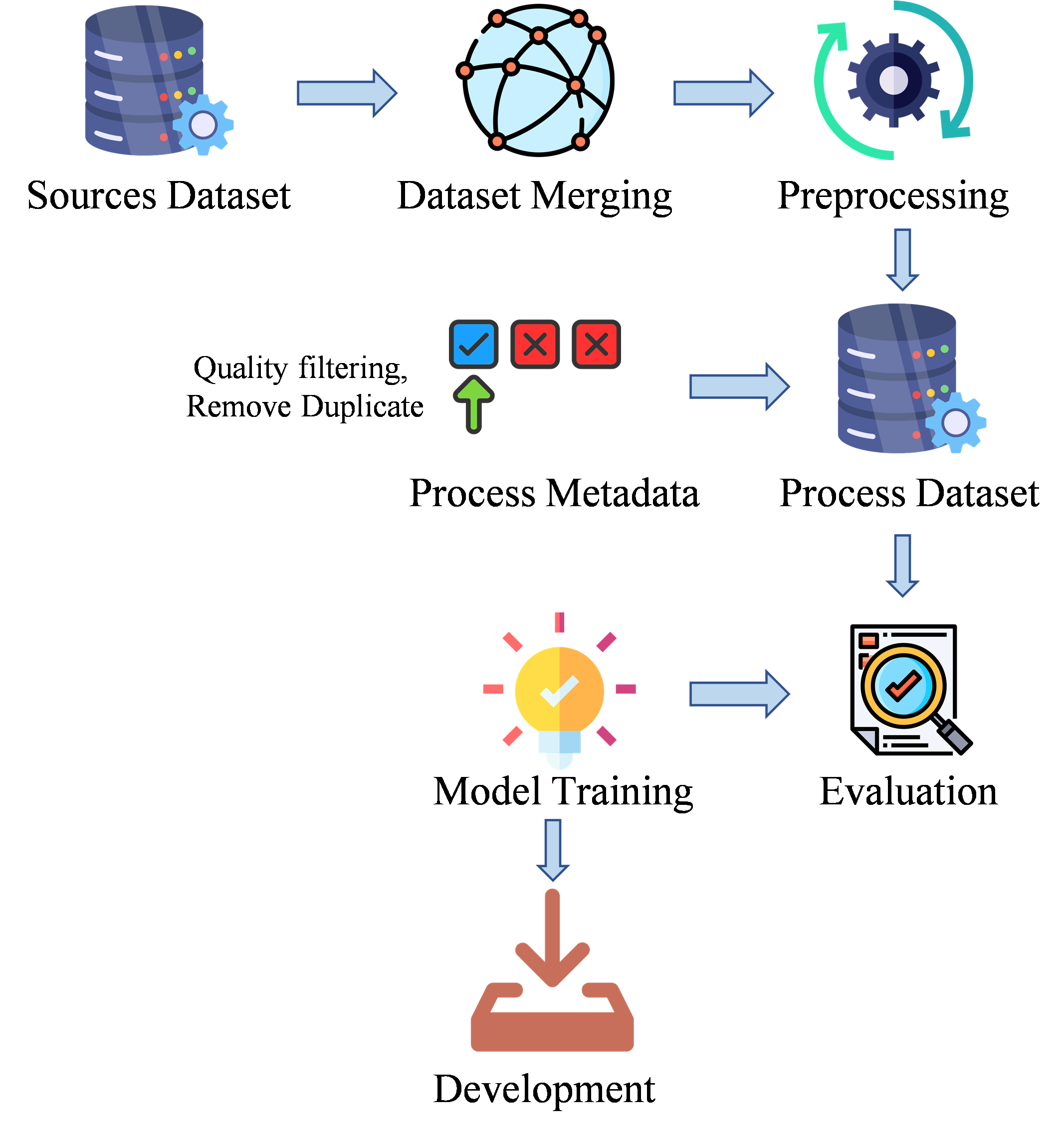}
    \caption{Workflow for creating GWD archive}
    \label{fig:workflow}
\end{figure}

Fig \ref{fig:workflow} represents the entire workflow of our study, from initial data gathering to final model construction, to provide a clear and thorough understanding of how the dataset was created. 

Fig \ref{fig:workflow} outlines the sequential steps of the research pipeline, including dataset collection, integration of multiple open source datasets, preprocessing procedures such as resizing, normalization, and duplicate removal, organization of annotations, and final preparation of the data for model training and evaluation. This visual overview improves clarity, supports reproducibility, and helps researchers better understand how each component contributes to building a reliable and robust dataset for waste classification and environmental monitoring applications.

\subsection{Data Source Selection} 
\begin{figure}[!h]
    \centering
    \includegraphics[width=0.9\linewidth]{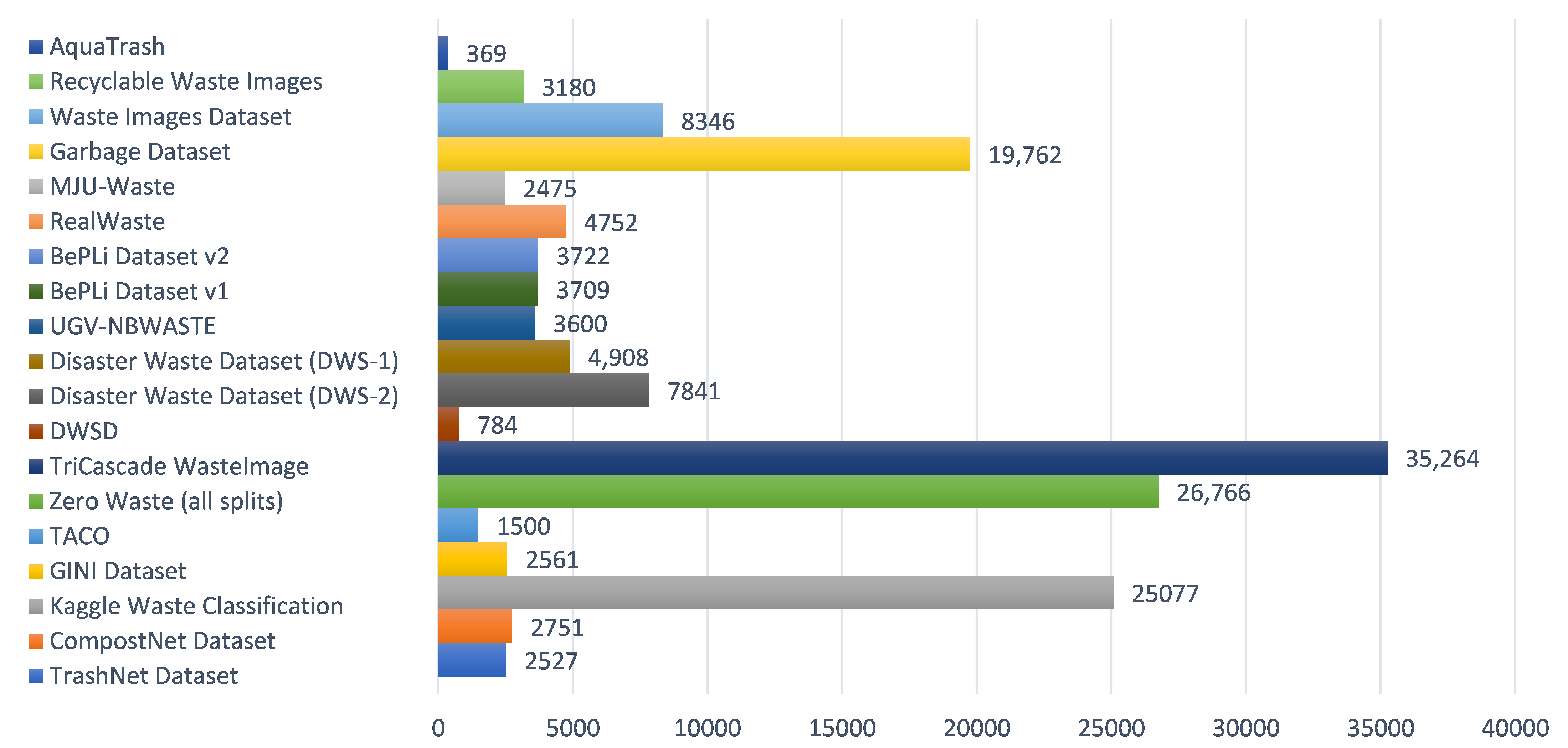}
    \caption{Waste classification datasets with their sample sizes}
    \label{fig:dataset-size}
\end{figure}

This novel GWD archive addresses the limitations of existing waste classification datasets by integrating a wide range of waste images from multiple open source collections into one comprehensive resource. Novel GWD archive is constructed using 19 publicly available datasets published between 2016 and 2025, resulting in a diverse and unified dataset for waste classification research. Selected datasets along with their sample size are represented in Fig \ref{fig:dataset-size}. Using multiple sources helped reduce bias, increase variability in visual conditions, and improve the overall robustness of the dataset.

\begin{figure}[!h]
    \centering
    \includegraphics[width=0.9\linewidth]{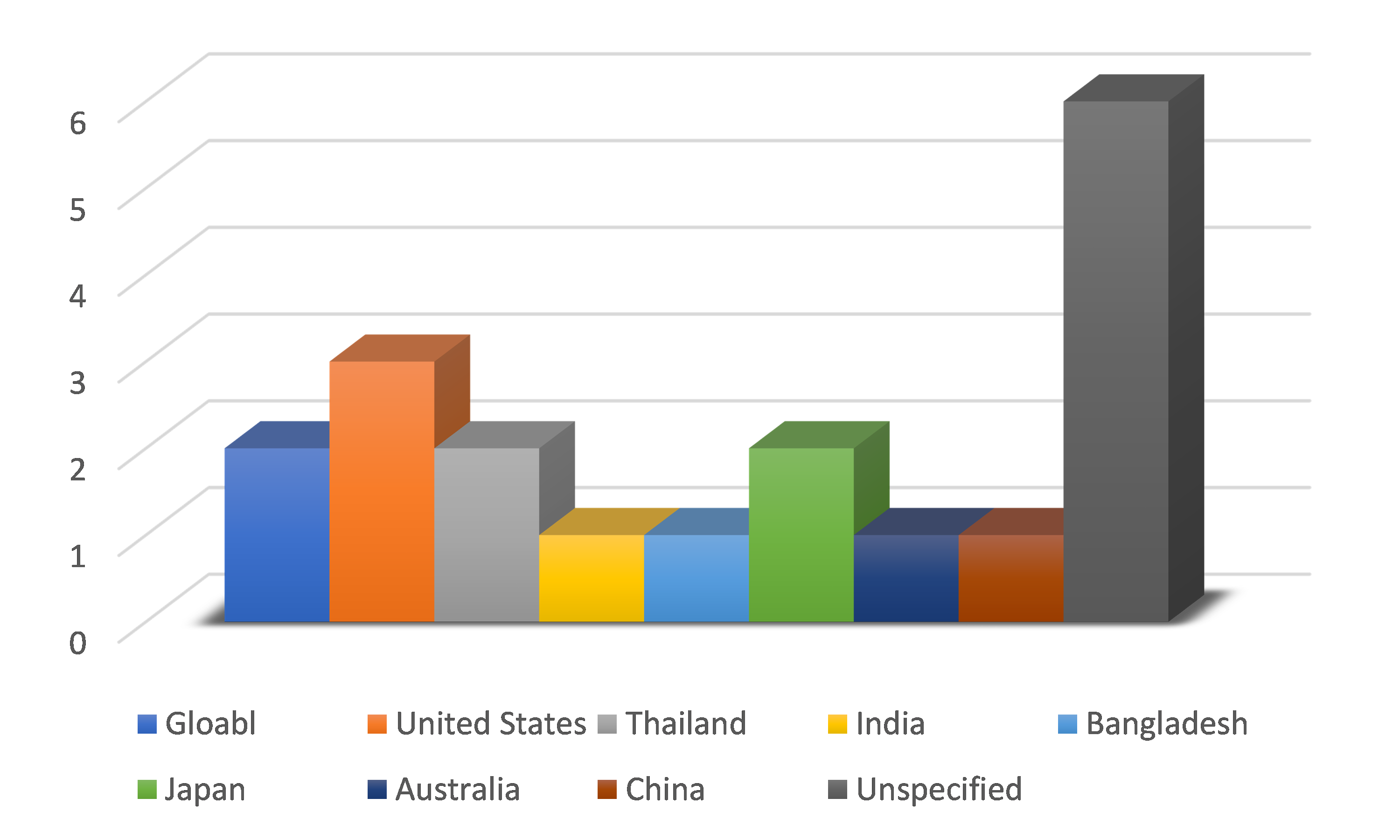}
    \caption{Geographic distribution of source datasets}
    \label{fig:geographic-distribution}
\end{figure}

The novel dataset reduces geographic bias and enhances cross regional generalization by uniformly integrating data from multiple regions, as shown in Fig \ref{fig:geographic-distribution}. By integrating data from different open source repositories, the final dataset benefits from a wider range of environments, object appearances, and capture conditions, which enhances its suitability for training and evaluating ML models.

This novel GWD archive obtained through a variety of methods, such as mobile based crowdsourced photos, field photos taken at waste collection locations, and controlled indoor captures. Every image was reorganized into a common directory hierarchy and annotation format to guarantee consistency. To capture both organized laboratory circumstances and unstructured real world waste environments, the novel dataset integrates these many data sources as shown in Table \ref{tab: collection methods}.

\begin{table}[!h]
\caption{Overview of data collection methods}\label{tab: collection methods}%
\begin{tabular}{@{} p{3cm} p{4cm} p{4cm} @{}}
\toprule
\textbf{Collection Method} & \textbf{Description} & \textbf{Tools/Platforms} \\
\midrule
Manual Collection & Field photography using smartphones, vehicle mounted devices, and vision sensors	& Smartphones, embedded cameras \\ 
Web Scraping & Automated download from online repositories & Google Images, Bing, Flickr \\ 
Crowdsourcing & Community submitted images through online platforms & Shared drives, GitHub, Flickr \\ 
\botrule
\end{tabular}
\end{table}

Table \ref{tab: collection methods} represent the variety of dataset collection methods employed in the development of the novel GWD archive. The majority of the datasets examined came from manual data collection, followed by web scraped and mixed source datasets. Model generalization is improved and environmental variety is increased by incorporating different acquisition mechanisms. 

Existing waste classification datasets differ significantly in how their images are collected, annotated, and licensed, which directly affects their applicability for large scale ML research. Most datasets, such as TrashNet \cite{Trashnet}, DSWD \cite{DWSD}, and RealWaste Dataset \cite{realwaste}, rely on manually captured images taken in controlled indoor environments using mobile phone cameras. These datasets typically contain clean, centered object images with uniform backgrounds, resulting in low environmental variability. In contrast, datasets like TACO \cite{taco} employ an internet scraping approach, collecting real world litter images from online platforms. This method increases diversity in scene context, background complexity, and lighting conditions but often introduces inconsistencies in image quality and class representation. A few other datasets use crowdsourced contributions, where volunteers upload images of waste items in natural settings, although such datasets remain relatively small and noisy. Table \ref{tab:collection-method} presents the unique collection methods and how many datasets used that method.
\begin{table}[!h]
\caption{Data acquisition strategies for constituent waste datasets}\label{tab:collection-method}%
\begin{tabular}{@{}lll@{}}
\toprule
Data Source & Acquisition Methods & Number of Datasets \\
\midrule
\cite{Trashnet, zerowaste, DWSD, DWS1, DWS2, UGV-NBWASTE, bepli_V1, BePLI_V2, realwaste, MJU_waste} & Manually collected  & 10 \\
\cite{waste_classification_data, GINI} & Web Scrapping                    & 2  \\ 
\cite{taco} &  Crowdsourcing        & 1  \\ 
\cite{compostnet} & Existing dataset + manually collected & 1  \\ 
\cite{aquatrash} & Based on a subset of the existing dataset & 1  \\ 
\cite{TriCasecade} & Combination of existing Datasets      & 1  \\ 
\cite{Garbage_dataset, WasteImagesdataset, RecyclableWaste} & Unspecified                           & 3  \\ 
\botrule
\end{tabular}
\end{table}
\subsection{Annotation methods}
\begin{figure} [!h]
    \centering
    \includegraphics[width=0.9\linewidth]{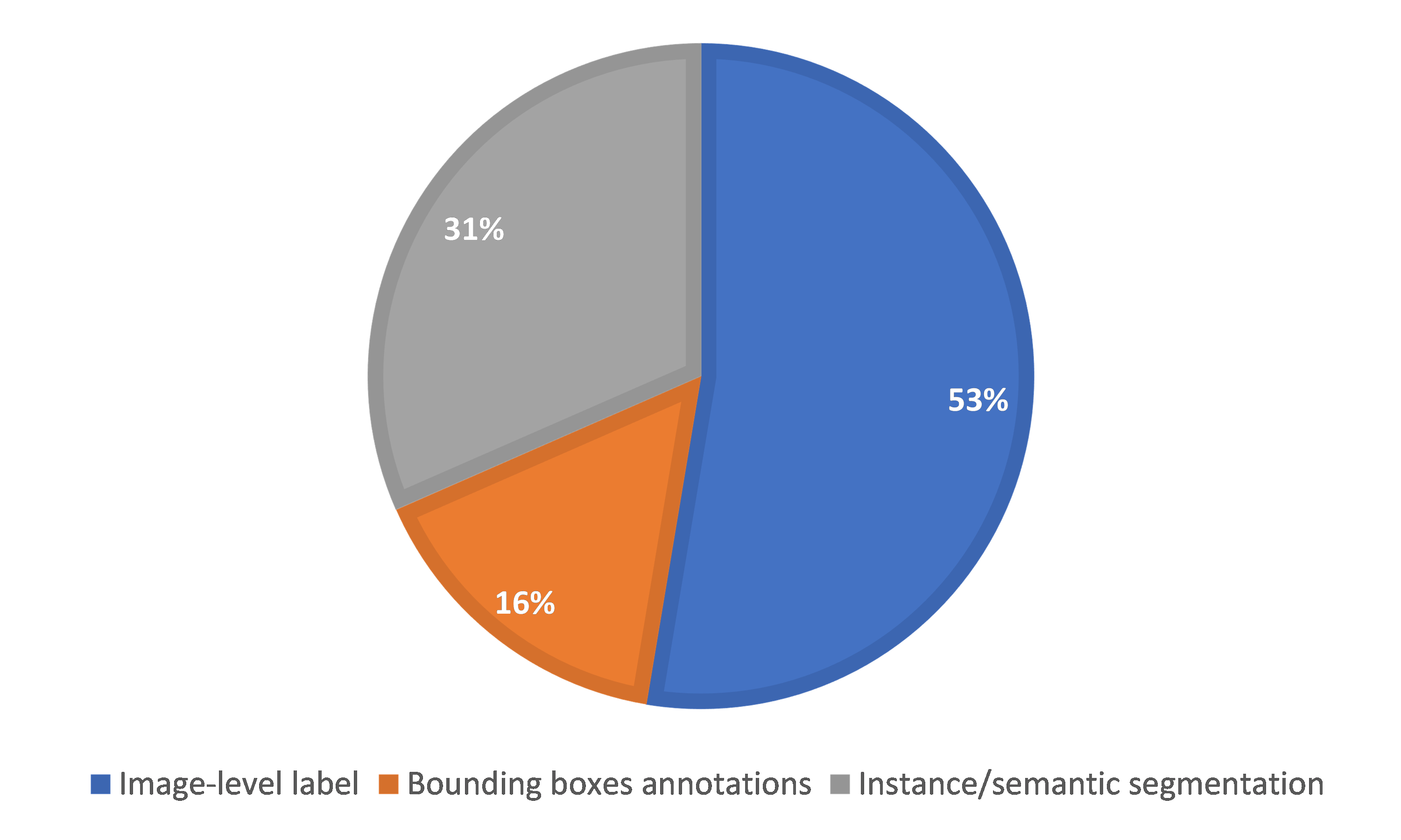}
    \caption{Annotation methods of source datasets}
    \label{fig: Annotation-methods}
\end{figure}

The datasets originally used different labeling schemes and directory structures as represent in the Fig \ref{fig: Annotation-methods}.
The Novel GWD archive employs a directory based (folder level) annotation strategy, which is a widely adopted and efficient approach for image classification tasks. In this method, each image is assigned a single class label based on the directory in which it is stored. The folder names correspond directly to waste categories and sub categories, ensuring a clear and consistent mapping between images and class labels.

To support both coarse and fine grained classification, the dataset is organized hierarchically. The top level directories represent main waste categories, while nested sub directories define specific waste types. This structure enables flexible model training at multiple semantic levels, allowing researchers to perform either broad category classification or detailed subclass recognition using the same dataset. Ambiguous or mislabeled samples were corrected or removed to maintain annotation reliability. Compared to bounding box or segmentation based annotation, this directory based labeling approach significantly reduces annotation complexity while remaining well suited for large scale classification and benchmarking tasks.

Overall, the folder based annotation scheme provides a lightweight, scalable, and reproducible labeling framework, making the dataset easily accessible for training and evaluating DL models in waste identification, recycling automation, and environmental monitoring applications.

The type and depth of annotation also vary across datasets. Earlier datasets such as CompostNet \cite{compostnet} and waste classification dataset \cite{waste_classification_data} typically provide single label classifications, assigning each image to a general waste class (e.g., plastic, metal, glass). More recent dataset \cite{TriCasecade} also uses single label or folder based labels. Particularly, TACO \cite{taco} includes detailed COCO style annotations, offering object level bounding boxes, instance segmentation masks, and multi label categories for complex scenes containing multiple waste items. Some datasets provide hierarchical labels, distinguishing between broad categories (e.g., recyclable vs. organic) and more specific subtypes. Table \ref{tab:annotation-types} describes the different annotation types utilized across the source datasets. However, most publicly available datasets lack standardized annotation protocols, resulting in inconsistencies across sources.
\begin{table}[!h]
\caption{Overview of annotation formats and labeling strategies with dataset counts}\label{tab:annotation-types}%
\begin{tabular}{@{}lll@{}}
\toprule
Data Source & Annotation Type & Number of Datasets \\
\midrule
\cite{Trashnet, compostnet, waste_classification_data, TriCasecade, DWS1, DWS2, realwaste, Garbage_dataset, WasteImagesdataset, RecyclableWaste} &  Image level label              & 10 \\ 
\cite{GINI, UGV-NBWASTE, aquatrash} &  Bounding boxes annotations     & 3  \\ 
\cite{taco, zerowaste, DWSD, bepli_V1, BePLI_V2, MJU_waste} & Instance/semantic segmentation & 6  \\ 
\botrule
\end{tabular}
\end{table}

\subsection{Licensing and Usage Rights}
\begin{figure} [!h]
    \centering
    \includegraphics[width=0.9\linewidth]{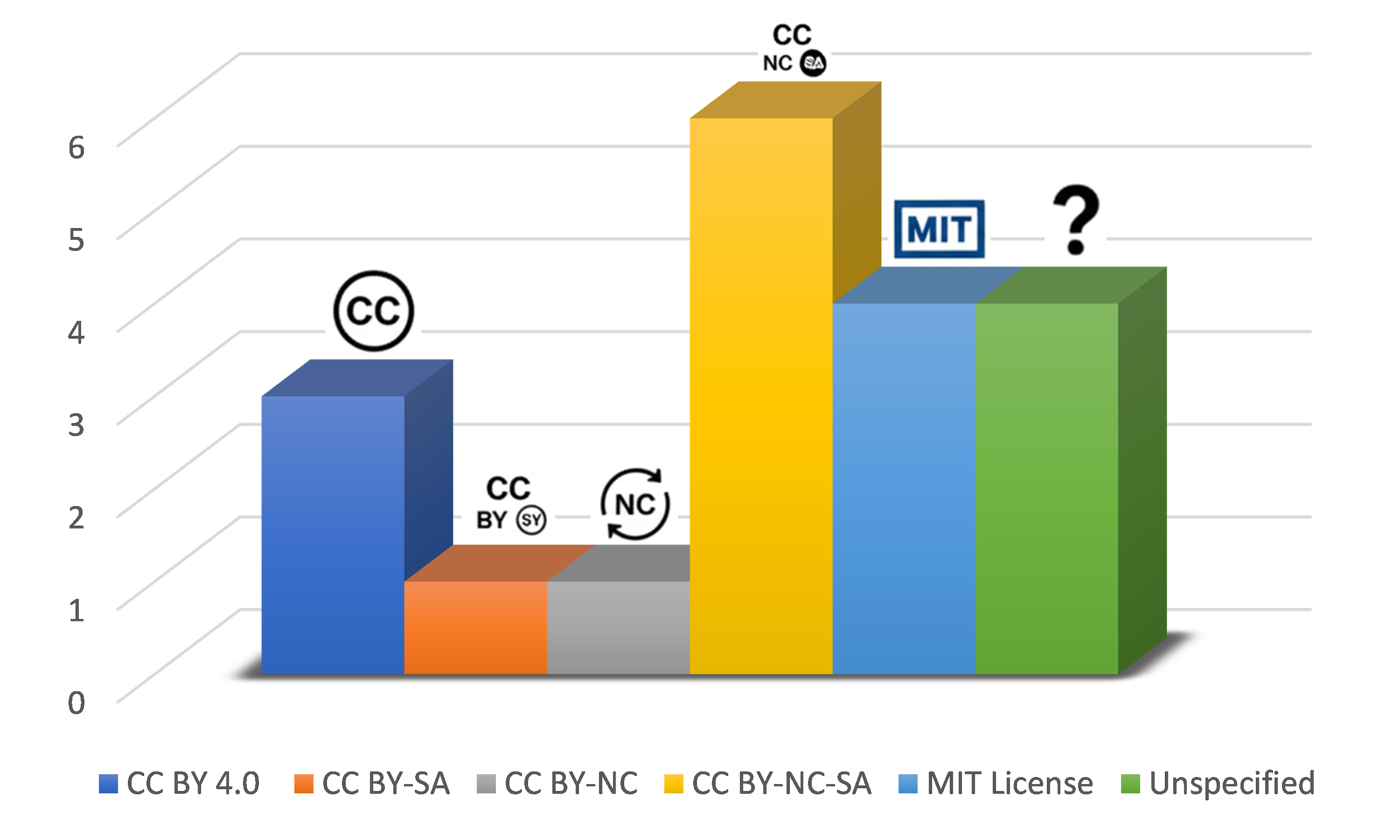}
    \caption{License types of source datasets}
    \label{fig: license-type}
\end{figure}

This Novel dataset is released under the Creative Commons Attribution 4.0 International License (CC BY 4.0). It permits unlimited use and sharing of the data, provided that appropriate credit is given to the original authors. Users are free to access and utilize the dataset with proper attribution; however, modification of the dataset is not permitted. Any commercial use remains subject to the licensing terms of the original data sources. The dataset is shared to support open, ethical, and reproducible research in the fields of waste classification and environmental monitoring. Fig \ref{fig: license-type} outlines the licenses retained for each original dataset included in this collection. When using the GWD archive, researchers are encouraged to acknowledge both the newly compiled dataset and the original data sources.

Licensing policies also differ among existing datasets. Many early datasets are released under Creative Commons licenses (e.g., CC BY 4.0), allowing academic use with attribution, while some restrict commercial reuse. Internet scraped datasets such as TACO operate under open research use licenses, permitting free access but requiring acknowledgment of original image sources. In contrast, several Kaggle datasets are distributed under Kaggle’s default open license, which allows research use but may limit redistribution of raw images. The lack of uniform licensing across datasets poses challenges when combining multiple sources into a single, unified dataset for large scale research. Table \ref{tab:license-type} represents the different licenses under which the datasets were released.

\begin{table}[!h]
\caption{Distribution of usage permissions for waste datasets}\label{tab:license-type}%
\begin{tabular}{@{}lll@{}}
\toprule
Data source & License Type & Number of Datasets \\
\midrule
\cite{taco, DWSD, UGV-NBWASTE} & CC BY 4.0   & 3 \\ 
\cite{waste_classification_data} &  CC BY-SA    & 1 \\ 
\cite{zerowaste} & CC BY-NC    & 1 \\ 
\cite{DWS1, DWS2, BePLI_V2, bepli_V1, realwaste, aquatrash} & CC BY-NC-SA & 6 \\ 
\cite{Trashnet, TriCasecade, MJU_waste, Garbage_dataset} & MIT License & 4 \\ 
\cite{compostnet, GINI, WasteImagesdataset, RecyclableWaste} & Unspecified & 4 \\ 
\botrule
\end{tabular}
\end{table}

\subsection{Data Preprocessing} 
All images in the dataset underwent a standardized preprocessing pipeline to ensure consistency across multiple source datasets. First, each image was resized to a fixed resolution (224 × 224) to unify varying dimensions and reduce computational complexity. Pixel values were then rescaled from the range [0, 255] to [0, 1], followed by channel wise normalization using the mean and standard deviation to stabilize training and improve model convergence. Since the dataset was compiled from several open source collections, class labels were cleaned and standardized into a unified taxonomy to remove naming inconsistencies and merge overlapping categories. Additionally, corrupted, duplicate, or extremely low quality images were removed to improve dataset reliability. Finally, the preprocessed dataset was divided into training (70\%), validation (10\%), and test (20\%) sets using a stratified split to maintain class balance across all subsets.

\section{Data Record}

The novel dataset consists of up of 89,807 imagess across 14 main categories, annotated with 68 distinct subclasses. The collection comprises photos taken in a variety of lighting situations (daylight, artificial, and low light), scenes (indoor and outdoor), and backgrounds (managed sets to naturally chaotic environments). The dataset is organized in a clear and consistent folder structure that facilitates ease of use across different ML frameworks. Each record in the dataset consists of an image file, its corresponding annotation, and a metadata entry describing contextual information about the sample.

\subsection{Repository Structure}

The dataset is organized into three primary partitions: training, validation, and testing. Each partition contains two subfolders—images/ and labels/—along with a metadata file describing the contents.

\subsection{Classes and Label Definition}
Novel GWD archive is organized using a hierarchical class taxonomy designed to reflect real world waste composition and recycling practices. The dataset contains 14 main waste categories, each further divided into 68 fine grained sub classes, enabling both coarse level and detailed waste recognition. This hierarchical structure supports flexible experimentation, allowing models to be trained either on high level material groups or on specific waste types.

\begin{figure}[!h]
    \centering
    \includegraphics[width=0.7\linewidth]{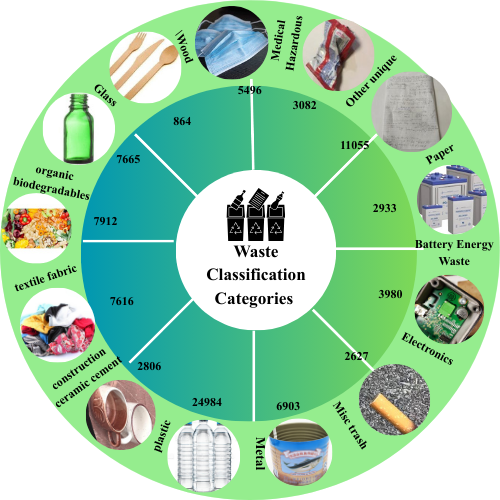}
    \caption{Distribution of main waste categories of GWD archive}
    \label{fig:waste sample}
\end{figure}

At the top level, the main categories represent broad waste classes such as plastics, paper and cardboard, metals, glass, organic waste, electronic waste, medical and hazardous items, textiles, construction debris, and other mixed materials. Each main category is subdivided into semantically meaningful sub classes (e.g., bottles, containers, packaging, food waste, clothing, cables, ceramics), capturing the visual and material diversity present in real waste streams. To provide a clear understanding of the dataset composition, a sample of each category is represent in Fig \ref{fig:waste sample}.

Fig \ref{fig:waste sample} illustrates the types of samples included in each main category, along with the corresponding sample size for each category.
Class labels are assigned using a directory based annotation scheme, where each folder name corresponds directly to a class label. This approach ensures a clear and unambiguous mapping between images and labels while maintaining compatibility with standard DL frameworks. The hierarchical folder structure further enables multi level classification without requiring additional annotation files.

\section{Conclusion}
The proposed Novel GWD archive offers a large scale, diverse, and hierarchically organized resource designed to overcome the limitations of existing small scale and domain specific datasets. By combining images from multiple openly available sources and applying rigorous preprocessing and standardization steps, the dataset achieves consistent labeling, enhanced image quality, and a balanced distribution across 14 main categories and 68 sub classes. This well structured composition makes the dataset highly suitable for training and evaluating ML models for applications such as automated waste sorting, environmental monitoring, and recycling optimization. Its public availability supports transparent benchmarking and fosters continued research in sustainable waste management technologies. Overall, the dataset represents a significant foundational contribution to the field and creates new opportunities for developing more accurate, scalable, and real world waste recognition systems.

\section*{Supplementary material}

To improve usability and reproducibility, class names follow a consistent naming convention, and each label is uniquely defined within the dataset taxonomy. The combination of broad and fine grained labels makes the dataset suitable for a wide range of applications, including waste material recognition, recycling automation, and environmental monitoring. Table \ref{tab:waste_dataset_breakdown} illustrate the compplete breakdown of novel dataest. 

\begin{longtable}{@{}lccr@{}}
\caption{Hierarchical distribution of primary waste categories and their corresponding sub category breakdown of GWD Archive}
\label{tab:waste_dataset_breakdown} \\
\toprule
\textbf{Main Category} & \textbf{Sub-Folder} & \textbf{Image Count} & \textbf{Category Total} \\
\midrule
\endfirsthead

\toprule
\textbf{Main Category} & \textbf{Sub-Folder} & \textbf{Image Count} & \textbf{Category Total} \\
\midrule
\endhead

\midrule

\endfoot

\bottomrule
\endlastfoot

\multirow{1}{*}{Battery\_Energy\_Waste} & battery & 2933 & \multirow{1}{*}{\textbf{2933}} \\ 
\midrule
\multirow{4}{*}{Construction\_Ceramic\_Cement} & cement\_based\_materials & 1010 & \multirow{4}{*}{\textbf{2806}} \\ 
 & ceramic\_materials & 552 & \\ 
 & construction\_scrap & 431 & \\ 
 & porcelin & 813 & \\ 
\midrule
\multirow{5}{*}{Electronic} & electrical\_cables & 553 & \multirow{5}{*}{\textbf{3980}} \\ 
 & electronic\_chips & 492 & \\ 
 & electronic\_components & 1578 & \\ 
 & laptops & 398 & \\ 
 & small\_appliances & 740 & \\ 
 & smartphones & 219 & \\ 
\midrule
\multirow{4}{*}{Glass} & brown\_glass & 607 & \multirow{4}{*}{\textbf{7665}} \\ 
 & glass & 3982 & \\ 
 & glass\_materials & 1672 & \\ 
 & green\_glass & 629 & \\ 
 & white\_glass & 775 & \\ 
\midrule
\multirow{6}{*}{Medical\_Hazardous} & bandai & 405 & \multirow{6}{*}{\textbf{5496}} \\ 
 & diaper & 778 & \\ 
 & gloves & 353 & \\ 
 & masks & 400 & \\ 
 & medicine\_and\_strip & 1307 & \\ 
 & pesticide\_bottle & 940 & \\ 
 & syringe & 405 & \\ 
 & thermometer & 908 & \\ 
\midrule
\multirow{5}{*}{Metal} & aluminium & 345 & \multirow{5}{*}{\textbf{6903}} \\ 
 & beverage\_cans & 1602 & \\ 
 & metal & 2630 & \\ 
 & metal\_containers & 437 & \\ 
 & metallic\_materials & 1489 & \\ 
 & spray\_cans & 400 & \\ 
\midrule
\multirow{2}{*}{Misc\_Trash} & cigarette\_butt & 97 & \multirow{2}{*}{\textbf{2627}} \\ 
 & garbage & 904 & \\ 
 & trash & 1626 & \\ 
\midrule
\multirow{4}{*}{Organic\_Biodegradable} & animal\_dead\_body & 220 & \multirow{4}{*}{\textbf{7912}} \\ 
 & compost & 177 & \\ 
 & foods & 5522 & \\ 
 & organic\_wastes & 1557 & \\ 
 & vegetation & 436 & \\ 
\midrule
\multirow{1}{*}{Others\_Unique} & others & 3082 & \multirow{1}{*}{\textbf{3082}} \\ 
\midrule
\multirow{5}{*}{Paper\_Cardboard} & cardboard & 4016 & \multirow{5}{*}{\textbf{11055}} \\ 
 & carton & 349 & \\ 
 & newspaper & 1120 & \\ 
 & paper\_cups & 511 & \\ 
 & papers & 4217 & \\ 
 & tetra\_pak & 842 & \\ 
\midrule
\multirow{15}{*}{Plastics} & box\_shaped\_case & 672 & \multirow{15}{*}{\textbf{24984}} \\ 
 & buoy & 1124 & \\ 
 & fishing\_net & 867 & \\ 
 & fragment & 3406 & \\ 
 & other\_bottle & 1103 & \\ 
 & other\_container & 747 & \\ 
 & other\_fishing\_gear & 549 & \\ 
 & other\_string & 513 & \\ 
 & pet\_bottle & 1433 & \\ 
 & plastic & 3869 & \\ 
 & plastic\_bag & 2007 & \\ 
 & plastic\_bottle & 755 & \\ 
 & plastic\_container & 464 & \\ 
 & plastic\_cups & 405 & \\ 
 & plastic\_litter & 3709 & \\ 
 & plastic\_materials & 1466 & \\ 
 & styrene\_foam & 1895 & \\ 
\midrule
\multirow{1}{*}{Rubber\_Polymer} & rope & 1027 & \multirow{1}{*}{\textbf{1884}} \\ 
 & rubber\_materials & 857 & \\ 
\midrule
\multirow{2}{*}{Textiles\_Fabric} & clothes & 5321 & \multirow{2}{*}{\textbf{7616}} \\ 
 & shoes & 1977 & \\ 
 & textile\_trash & 318 & \\ 
\midrule
\multirow{1}{*}{Wood} & wooden\_materials & 864 & \multirow{1}{*}{\textbf{864}} \\ 

\end{longtable}

\section*{Declarations}
\begin{itemize}
\item \textbf{Ethics approval and consent to participate.}
\item \textbf{Funding.} No funding
\item \textbf{Declaration of competing interest.} The authors declare that they have no known competing financial interests or personal relationships that could have appeared to influence the work reported in this paper.
\item \textbf{Consent for publication.} Not applicable
\item \textbf{Data availability.} Data is publicly available at: \href{https://cloud.dfki.de/owncloud/index.php/s/656dTBGXKQBaPPB}{Dataset}
\item \textbf{CRediT authorship contribution statement.} Misbah Ijaz \& Saif Ur Rehman Khan: Conceptualization, Data curation, Methodology, Software, Validation, Writing original draft \& Formal analysis. Muhammed Nabeel Asim, Sebastian Vollmer \& Andreas Dengel: Conceptualization, Funding acquisition, Review. Abd Ur Rehman \& Tayyaba Asif: Supervision, review \& editing.
\end{itemize}

\bibliography{sn-bibliography.bib}

@article{wastegeneration,
title = {Waste generation, waste disposal and policy effectiveness: Evidence on decoupling from the European Union},
journal = {Resources, Conservation and Recycling},
volume = {52},
number = {10},
pages = {1221-1234},
year = {2008},
issn = {0921-3449},
doi = {https://doi.org/10.1016/j.resconrec.2008.07.003},
url = {https://www.sciencedirect.com/science/article/pii/S0921344908001079},
author = {Massimiliano Mazzanti and Roberto Zoboli}
}

@incollection{Chapter_33,
title = {Chapter 33 - Waste management practices in the developing nations: challenges and opportunities},
editor = {Pardeep Singh and Pramit Verma and Rishikesh Singh and Arif Ahamad and André C.S. Batalhão},
booktitle = {Waste Management and Resource Recycling in the Developing World},
publisher = {Elsevier},
pages = {773-797},
year = {2023},
isbn = {978-0-323-90463-6},
doi = {https://doi.org/10.1016/B978-0-323-90463-6.00017-8},
url = {https://www.sciencedirect.com/science/article/pii/B9780323904636000178},
author = {Tanu Kumari and Akhilesh Singh Raghubanshi}
}

@INPROCEEDINGS{smartwaste,
  author={Umer, Hamid and Ahmed, Awais and Ali, Farman and Sarmad Ali, Syed and Ali Khan, Munim},
  booktitle={2022 Mohammad Ali Jinnah University International Conference on Computing (MAJICC)}, 
  title={A Systematic Literature Review on Smart Waste Management Using Machine Learning}, 
  year={2022},
  volume={},
  number={},
  pages={1-9},
  doi={10.1109/MAJICC56935.2022.9994104}}

@article{Trashnet,
  title={Classification of trash for recyclability status},
  author={Yang, Mindy and Thung, Gary},
  journal={CS229 project report},
  volume={2016},
  number={1},
  pages={3},
  year={2016}
}

@inproceedings{trashbox,
  title={Trashbox: trash detection and classification using quantum transfer learning},
  author={Kumsetty, Nikhil Venkat and Nekkare, Amith Bhat and Kamath, Sowmya and others},
  booktitle={2022 31st Conference of Open Innovations Association (FRUCT)},
  pages={125--130},
  year={2022},
  organization={IEEE}
}

@inproceedings{WaDaBa,
  title={PET waste classification method and plastic waste DataBase-WaDaBa},
  author={Bobulski, Janusz and Piatkowski, Jacek},
  booktitle={International conference on image processing and communications},
  pages={57--64},
  year={2017},
  organization={Springer}
}

@article{wasteclassification,
  title={Solid waste image classification using deep convolutional neural network},
  author={Nnamoko, Nonso and Barrowclough, Joseph and Procter, Jack},
  journal={Infrastructures},
  volume={7},
  number={4},
  pages={47},
  year={2022},
  publisher={MDPI}
}

@article{taco,
  title={Taco: Trash annotations in context for litter detection},
  author={Proen{\c{c}}a, Pedro F and Sim{\~o}es, Pedro},
  journal={arXiv preprint arXiv:2003.06975},
  year={2020}
}

@inproceedings{GINI,
  title={Spotgarbage: smartphone app to detect garbage using deep learning},
  author={Mittal, Gaurav and Yagnik, Kaushal B and Garg, Mohit and Krishnan, Narayanan C},
  booktitle={Proceedings of the 2016 ACM international joint conference on pervasive and ubiquitous computing},
  pages={940--945},
  year={2016}
}

@article{aquatrash,
  title={AquaVision: Automating the detection of waste in water bodies using deep transfer learning},
  author={Panwar, Harsh and Gupta, PK and Siddiqui, Mohammad Khubeb and Morales-Menendez, Ruben and Bhardwaj, Prakhar and Sharma, Sudhansh and Sarker, Iqbal H},
  journal={Case Studies in Chemical and Environmental Engineering},
  volume={2},
  pages={100026},
  year={2020},
  publisher={Elsevier}
}

@article{DWSD,
  title={DWSD: Dense waste segmentation dataset},
  author={Ali, Asfak and Acharjee, Suvojit and Sk, Md Manarul and Alharthi, Salman Z and Chaudhuri, Sheli Sinha and Akhunzada, Adnan},
  journal={Data in Brief},
  volume={59},
  pages={111340},
  year={2025},
  publisher={Elsevier}
}

@article{fotovvatikhah2025systematic,
  title={A Systematic Review of AI-Based Techniques for Automated Waste Classification},
  author={Fotovvatikhah, Farnaz and Ahmedy, Ismail and Noor, Rafidah Md and Munir, Muhammad Umair},
  journal={Sensors (Basel, Switzerland)},
  volume={25},
  number={10},
  pages={3181},
  year={2025}
}

@article{chomicki2025assessing,
  title={Assessing the impact of dataset quality on the performance of artificial intelligence models in automatic waste classification},
  author={Chomicki, Adam and W{\'o}jcik, Filip and Dudycz, Helena},
  journal={Procedia Computer Science},
  volume={270},
  pages={1061--1070},
  year={2025},
  publisher={Elsevier}
}

@article{bepli_V1,
  title={BePLi Dataset v1: Beach Plastic Litter Dataset version 1 for instance segmentation of beach plastic litter},
  author={Hidaka, Mitsuko and Murakami, Koshiro and Koshidawa, Kenta and Kawahara, Shintaro and Sugiyama, Daisuke and Kako, Shin'ichiro and Matsuoka, Daisuke},
  journal={Data in Brief},
  volume={48},
  pages={109176},
  year={2023},
  publisher={Elsevier}
}

@article{BePLI_V2,
  title={Updating “BePLi Dataset v1: Beach Plastic Litter Dataset version 1, for instance segmentation of beach plastic litter” with 13 object classes},
  author={Hidaka, Mitsuko and Murakami, Koshiro and Kawahara, Shintaro and Nakagawa, Yujin and Sugiyama, Daisuke and Kako, Shin’ichiro and Matsuoka, Daisuke},
  journal={Data in Brief},
  pages={111867},
  year={2025},
  publisher={Elsevier}
}

@inproceedings{compostnet,
  title={Compostnet: An image classifier for meal waste},
  author={Frost, Sarah and Tor, Bryan and Agrawal, Rakshit and Forbes, Angus G},
  booktitle={2019 IEEE Global Humanitarian Technology Conference (GHTC)},
  pages={1--4},
  year={2019},
  organization={IEEE}
}

@article{Garbage_dataset,
  title={Managing household waste through transfer learning},
  author={Kunwar, Suman},
  journal={arXiv preprint arXiv:2402.09437},
  year={2024}
}

@misc{waste_classification_data,
  title        = {Waste Classification Data},
  howpublished = {Kaggle},
  note         = {Available at: https://www.kaggle.com/datasets/techsash/waste-classification-data (accessed on 31 July 2024)},
  author       = {Techsash}
}

@article{MJU_waste,
  title={A multi-level approach to waste object segmentation},
  author={Wang, Tao and Cai, Yuanzheng and Liang, Lingyu and Ye, Dongyi},
  journal={Sensors},
  volume={20},
  number={14},
  pages={3816},
  year={2020},
  publisher={MDPI}
}

@article{realwaste,
  title={RealWaste: a novel real-life data set for landfill waste classification using deep learning},
  author={Single, Sam and Iranmanesh, Saeid and Raad, Raad},
  journal={Information},
  volume={14},
  number={12},
  pages={633},
  year={2023},
  publisher={MDPI}
}

@article{UGV-NBWASTE,
  title={UGV-NBWASTE: An oriented dataset for non-biodegradable waste in Bangladesh},
  author={Isalm, Md Riadul and Mahabub, Nabil Bin and Rafi, Md Jubayar Alam and Roy, Pronoy Kanti and Roy, Turjo and Islam, Md Tariqul and Razzak, Md Abdur},
  journal={Data in Brief},
  volume={60},
  pages={111559},
  year={2025},
  publisher={Elsevier}
}

@inproceedings{zerowaste,
  title={Zerowaste dataset: Towards deformable object segmentation in cluttered scenes},
  author={Bashkirova, Dina and Abdelfattah, Mohamed and Zhu, Ziliang and Akl, James and Alladkani, Fadi and Hu, Ping and Ablavsky, Vitaly and Calli, Berk and Bargal, Sarah Adel and Saenko, Kate},
  booktitle={Proceedings of the IEEE/CVF conference on computer vision and pattern recognition},
  pages={21147--21157},
  year={2022}
}

@article{TriCasecade,
  title={An automated waste classification system using deep learning techniques: Toward efficient waste recycling and environmental sustainability},
  author={Nahiduzzaman, Md and Ahamed, Md Faysal and Naznine, Mansura and Karim, Md Jawadul and Kibria, Hafsa Binte and Ayari, Mohamed Arselene and Khandakar, Amith and Ashraf, Azad and Ahsan, Mominul and Haider, Julfikar},
  journal={Knowledge-Based Systems},
  volume={310},
  pages={113028},
  year={2025},
  publisher={Elsevier}
}

@misc{DWS1,
  author = {{Engineering UBU}},
  title = {{Disaster Waste Dataset}},
  howpublished = {Kaggle},
  year = {2023},
  url = {https://www.kaggle.com/datasets/engineeringubu/disaster-waste-dataset},
  note = {Accessed: 2025-12-13}
}

@misc{DWS2,
  author = {{Engineering UBU}},
  title = {{Disaster Waste Dataset: DWS-2 Dataset Subset}},
  howpublished = {Kaggle},
  year = {2023},
  url = {https://www.kaggle.com/datasets/engineeringubu/disaster-waste-dataset/?select=DWS-2+dataset},
  note = {Accessed: 2025-12-13}
}

@misc{WasteImagesdataset,
  author = {Ouedraogo, Angelika Sita},
  title = {{Waste Images}},
  howpublished = {Kaggle},
  year = {2023},
  url = {https://www.kaggle.com/datasets/angelikasita/waste-images},
  note = {Accessed: 2025-12-13}
}

@misc{RecyclableWaste,
  author = {Dragicevic, Sanja},
  title = {{Recyclable Waste Images}},
  howpublished = {Kaggle},
  year = {2021},
  url = {https://www.kaggle.com/datasets/sanjadrag24/recyclable-waste-images},
  note = {Accessed: 15-December-2025}
}

\end{document}